\documentclass[10pt,twocolumn,letterpaper]{article}

\usepackage{cvpr}      

\usepackage{amsmath,amssymb,amsfonts}
\usepackage{graphicx}
\usepackage{textcomp}
\usepackage{algorithm}
\usepackage{algpseudocode}
\usepackage{xcolor}
\usepackage{url}
\def\BibTeX{{\rm B\kern-.05em{\sc i\kern-.025em b}\kern-.08em
    T\kern-.1667em\lower.7ex\hbox{E}\kern-.125emX}}

\definecolor{cvprblue}{rgb}{0.21,0.49,0.74}
\usepackage[pagebackref,breaklinks,colorlinks,allcolors=cvprblue]{hyperref}

\def\paperID{*****} 
\def\confName{CVPR}
\def\confYear{2025}

\title{Finding the Reflection Point: Unpadding Images to Remove Data Augmentation Artifacts in Large Open Source Image Datasets for Machine Learning
}

\author{Lucas Choi\\
\textit{Archbishop Mitty}\\
\and
Ross Greer\\
\textit{University of California, Merced}\\
}

\begin{document}
\maketitle

\begin{abstract}
In this paper, we address a novel image restoration problem relevant to machine learning dataset curation: the detection and removal of noisy mirrored padding artifacts. While data augmentation techniques like padding are necessary for standardizing image dimensions, they can introduce artifacts that degrade model evaluation when datasets are repurposed across domains. We propose a systematic algorithm to precisely delineate the reflection boundary through a minimum mean squared error approach with thresholding and remove reflective padding. 
Our method effectively identifies the transition between authentic content and its mirrored counterpart, even in the presence of compression or interpolation noise. We demonstrate our algorithm's efficacy on the SHEL5k dataset, showing significant performance improvements in zero-shot object detection tasks using OWLv2, with average precision increasing from 0.47 to 0.61 for hard hat detection and from 0.68 to 0.73 for person detection. By addressing annotation inconsistencies and distorted objects in padded regions, our approach enhances dataset integrity, enabling more reliable model evaluation across computer vision tasks.

\end{abstract}


\section{Introduction}


Data augmentation is a fundamental task in data preprocessing and training for deep learning tasks. However, when repurposing data between learning tasks or domains, images altered by task-specific augmentations are not always desired. Therefore, to recover the raw data when only an altered form is available, image restoration becomes a necessary computer vision subtask. Here, we introduce a novel and niche problem in image restoration, which can be introduced through errors in dataset curation during machine learning: the detection and removal of noisy mirrored regions. In mass image data collection or curation, especially when images may come from different cameras or feature cropped regions of interest of varying sizes, image dimensions may be inconsistent. However, many deep learning architectures require images of fixed input size. This is often remedied by resizing or padding; it is the padding case that we handle in this research. 

\begin{figure}
    \centering
\includegraphics[width=0.39\linewidth]{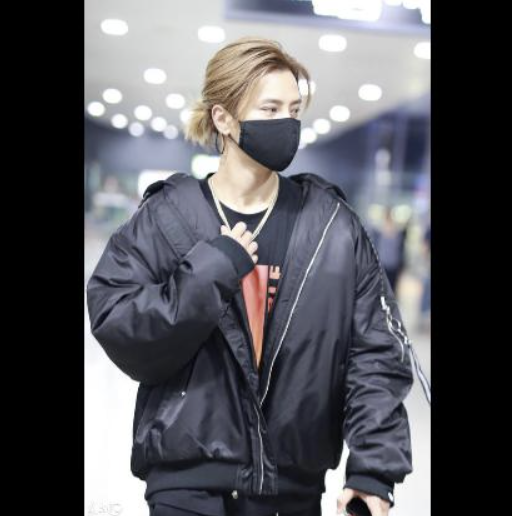}
\includegraphics[width=0.405\linewidth]{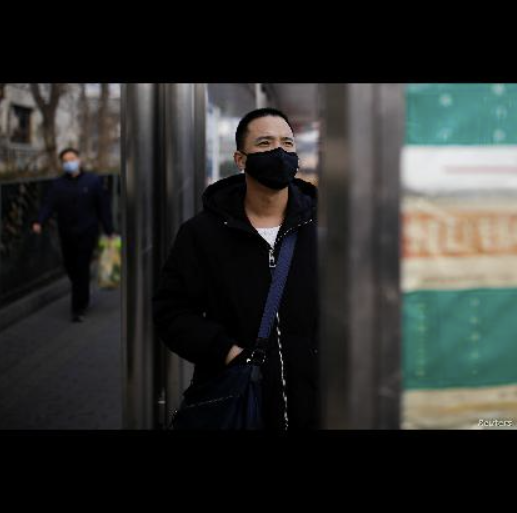} \\
\includegraphics[width=0.312\linewidth]{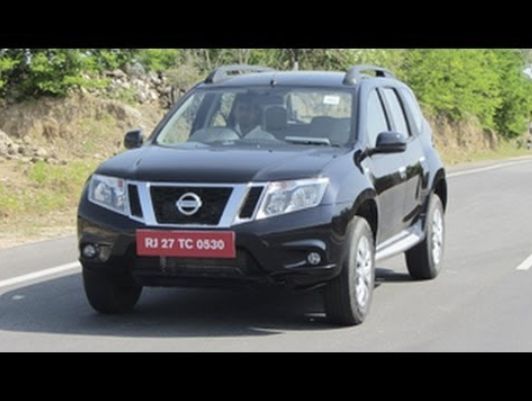}
\includegraphics[width=0.235\linewidth]{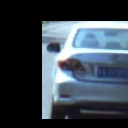}
\includegraphics[width=0.235\linewidth]{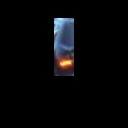} \\
\includegraphics[width=0.4\linewidth]{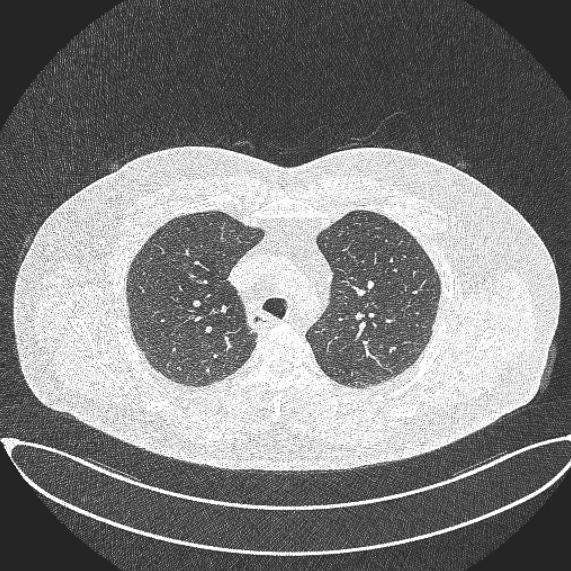}
\includegraphics[width=0.4\linewidth]{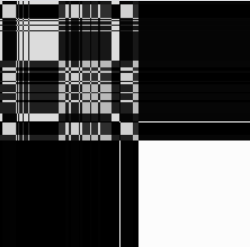} \\
\includegraphics[width=.196\textwidth]{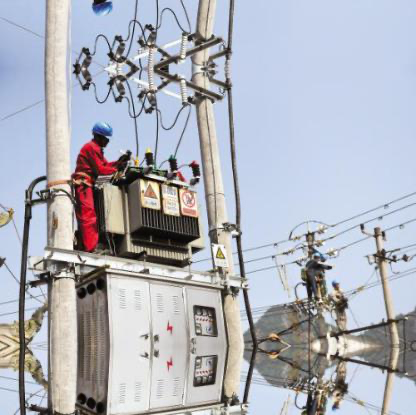} 
\includegraphics[width=.196\textwidth]{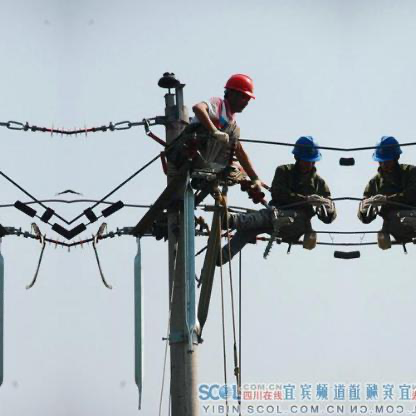} 

\caption{Example padded images from public datasets. Described from left to right and top to bottom: Zhang's Face Mask Dataset \cite{maskdataset}, Gohil's License Plate Dataset \cite{licenseplatedataset}, and the LISA Lights Dataset \cite{greer2024patterns, keskar2025lights}  contain zero-padding on vertical and horizontal edges. Das's IQ-OTH/NCCD Augmented Lung Cancer Dataset \cite{subhajeet_das_2025} contains radial zero padding. Fernandez's Markov Transition Field Images of Heart Beats \cite{heartdataset}, an image-based derivation of \cite{mark1982annotated}, contains unique padding for transition states beyond the areas of interest. Sample images from the SHEL5k dataset (bottom row), demonstrate our primary interest in this research, addressing the issue of noisy mirrored padding on the top and right for the first image, and the left and right for the second.}
    \label{fig:examples}
\end{figure}

As an alternative to padding at training time, machine learning practitioners may choose to pad their dataset prior to training as a precomputation step, which is also helpful in reducing repeated computations during multiple epochs of training or isolating features of interest for learning. Examples of such public dataset artifacts are shown in Figure \ref{fig:examples}.  However, if this padded dataset is saved and publicized instead of the original images, the artifacts it contains can lead to problems in evaluation, especially when the data used is transferred to other tasks. With large and redundant data volumes, the padding may be fine for training, but during evaluation, the presence of padding -- particularly symmetric or reflective padding --  can simultaneously create realistic objects or patterns that should be recognized but are left out of annotation, and unrealistic objects or patterns that should not be recognized and actually distort the meaning of the original object. In both cases, for symmetrically-padded images, using annotations centered on object detection of the original objects will lead to misleading performance evaluation. A solution we propose in this research is \textit{image unpadding}, where the padding on an image can be removed to restore an original image.

In cases of true zero-padding, the problem is trivial. When it comes to symmetric padding, further trivial methods, such as iterating through columns of pixels and identifying a consecutive repeating section in reverse, are only feasible when the image is saved without compression or resizing, as these processes can introduce interpolation noise. The presence of noise on either or both sides of a reflection makes detecting the boundary of an artificially-mirrored region non-trivial. Reflection removal requires a precise delineation of the mirrored boundary to separate authentic scene content from its redundant and often non-naturalistic counterpart. 


In this research, we propose a systematic approach to identify the reflection boundary, accurately localize the mirrored region, and remove redundant, inaccurate, or non-naturalistic information. We then demonstrate the effectiveness of this unpadding on inference and evaluation of a zero-shot object detection task.


\section{Related Research}
The necessity of image restoration, specifically unpadding mirrored regions, arises from the advent of image augmentation due to the necessity in variability and quantity of data with deep learning. To our knowledge, we are the first to address this issue and include further motivation in this section.

Having large volumes of high-quality data is paramount to the training of neural networks; for most tasks, the best-performing models are those that train on the greatest volume of data \cite{sun2017revisiting, geirhos2022bittersweet, klugscaling, schulz2020different}. In current tasks, this volume has increased to ``internet-scale" amounts of images, which can be thought of (in some regard) as a `dataset augmentation' by grabbing as much data as possible, even if from different sources, enabling transfer learning, foundation model learning, and more. This is a parallel path to standard data augmentation, where a dataset itself is used as the basis for generating different variations or augmentations on image samples. For example, a survey by Khalifa et al. \cite{khalifa2022comprehensive} describes the benefits that data augmentation contributes to deep-learning models. They note that image augmentation overcomes data scarcity, with fields such as medical imaging lacking sufficient labeled data \cite{chlap2021review}. Additionally, compared to collecting and labeling new data, augmentation provides a cost-effective alternative by transforming existing data into new samples. Augmentation techniques help reduce overfitting and ensure models achieve higher accuracy during testing by diversifying training samples \cite{shorten2019survey} and allowing for model generalizability \cite{xu2023comprehensive, nagaraju2022performance}. Finally, augmentation allows for more control over dataset characteristics, helping to balance class distributions and improve performance on imbalanced datasets \cite{mikolajczyk2018data}. 



Due to these advantages of both data and dataset augmentation, their usage is becoming prevalent. However, when altered datasets such as those exemplified in Figure \ref{fig:examples} are released without the original naturalistic data, it is difficult for future researchers to repurpose and apply the dataset for their own training, evaluation, or dataset augmentation. Aligned with these concerns, these augmentation or padding artifacts can be detrimental to neural network training and validation \cite{elgendi2021effectiveness}, especially as augmented data does not accurately represent real-world data, further emphasizing the necessity of naturalistic original data. 


\section{Algorithm for Image Unpadding}

Our algorithm for detecting the padded, mirrored area in an image is described in this section. We will explain the algorithm for just one side of the image (in this case, top), but we note that the algorithm should be applied to all four sides of the image. 

We first create a variable for the dividing line, which iterates from the top to the middle of the image. At each position of the dividing line, we crop from the top of the image to the line as well as a section of the image with the same area as the crop right below the dividing line. We then mirror the first cropped image over the x-axis and obtain the mean squared error (MSE), calculated as \[
\text{MSE} = \frac{1}{n} \sum_{i=1}^{n} (y_i - \hat{y}_i)^2
\] of these two cropped image segments. The line where the minimum MSE is located is then estimated to be the line between the padded area and the raw image, as the two cropped regions are most similar. 

However, it is also noted that the given images do not always have mirrored padding. To detect these instances and ensure that we do not necessarily crop parts of the raw image, we determine a threshold for the MSE, where if the MSE is greater than the threshold, the obtained dividing line is disregarded. This is based on the assumption that images without reflected padding should have a higher MSE since the content of the cropped images across the dividing line is not identical. 

When iterating the dividing line, we initially set it at some offset value from the boundary. This is since the portion of the image near the boundary sometimes has little to no difference in pixel values due to being part of the ground or sky. To clearly distinguish between unaltered images and padded images, the dividing-line offset ensures that MSE calculations focus on image regions where variability begins to appear. Without an offset, the minimum MSEs from unaltered images and padded images can be very low and overlapping in range due to cropping small sections of the image with low variability. By creating a clearer differentiation in the MSEs, we can better choose an MSE-threshold for unaltered and padded images. In the case of zero-padding, starting at the border would result in an extremely low MSE since the sections within the padding are identical. Therefore, it is necessary to start sufficiently positioned within the padding to increase the MSE. 
A pseudocode version of our algorithm is provided in Algorithm \ref{alg:mirror_detection}.

\begin{algorithm}
\caption{Detect Padded Mirrored Area at the Top}
\label{alg:mirror_detection}
\begin{algorithmic}
    \State \textbf{Input:} Image $I$, Threshold $\tau$, Offset $O$
    \State \textbf{Output:} Dividing line $L^*$ or border
    \State $MSE_{\min} \gets \infty$, $L^* \gets 0$
\For{$L \gets O$ \textbf{to} $\frac{\text{height}(I)}{2}$}

        \State $I_{\text{top}} \gets I[0:L]$
        \State $I_{\text{bot}} \gets I[L:2L]$
        \State $I_{\text{top}} \gets \text{mirror}(I_{\text{top}})$
        \State $MSE(L) \gets \text{MSE}(I_{\text{top}}, I_{\text{bot}})$
        \If{$MSE(L) < MSE_{\min}$}
            \State $MSE_{\min} \gets MSE(L)$
            \State $L^* \gets L$
        \EndIf
    \EndFor
    \If{$MSE_{\min} > \tau$}
        \State \textbf{Return} border
    \Else
        \State \textbf{Return} $L^*$
    \EndIf
\end{algorithmic}
\end{algorithm}



\subsection{Threshold Selection Algorithm}
We describe and compare various methods to obtain a threshold by labeling a small training set of images randomly selected from the dataset, which may or may not contain padding. The first method is iteration through threshold values and counting the number of dividing lines that are correctly estimated (within a pixel of tolerance). From this, an estimate of precision and recall can be formed for each threshold value, and the optimal threshold parameter for differentiating MSEs can be selected. As with most hyperparameter tuning, this result can be refined by repeating this process with smaller iteration step sizes around the previously obtained threshold.


The second method we employ is a variant of Otsu's thresholding method \cite{otsu1975threshold}, using MSEs of a training set rather than image intensity values. This involves taking the MSEs, normalizing them to a scale of [0, 255], and finding the Otsu's threshold of the normalized MSEs. Otsu’s method selects the threshold \( t^* \) that maximizes the between-class variance:
\[
\sigma_B^2(t) = \omega_1(t) \omega_2(t) (\mu_1(t) - \mu_2(t))^2
\]
where \( t \) is the threshold value, \( \omega_1 \) and \( \omega_2 \) are the class probabilities, and \( \mu_1 \) and \( \mu_2 \) are the class means. This gives the threshold to divide the lower and upper portions of the MSEs, which corresponds to the padded and unpadded images.



\section{Experimental Evaluation}

\subsection{Dataset}
To apply this reflected padding detection algorithm, we use an image dataset in a construction setting, namely the SHEL5k \cite{otgonbold2022shel5k}. This dataset has 5,000 images with mirrored padding of either the top and bottom or the left and right of each image, as shown in Figure \ref{fig:examples}. This is likely created by the creators of the dataset through image augmentation to resize the images. 

\begin{figure}
    \centering
    \includegraphics[width=.366\textwidth]{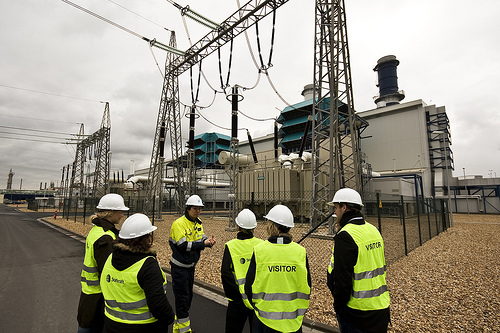} 
    \caption{Sample control image from the Hard Hat Workers Dataset.}
    \label{fig:hardhatexample}
\end{figure}

Additionally, we utilize 210 control images without any augmentation, taken from similar settings in the Hard Hat Workers Dataset \cite{hard-hat-workers_dataset}, to discern the threshold for the MSE, with an example shown in Figure \ref{fig:hardhatexample}.

\subsection{Threshold Selection}

To calculate the MSE threshold for the dividing line of the padding, we first sample 400 training images from the 5,000 images of the SHEL5k dataset \cite{otgonbold2022shel5k}, selecting 200 with padding and 200 without for training balance. We compute and store the minimal MSEs of each image according to the unpadding algorithm presented in Section 3. 

\begin{figure}
    \centering
    \includegraphics[width=.48\textwidth]{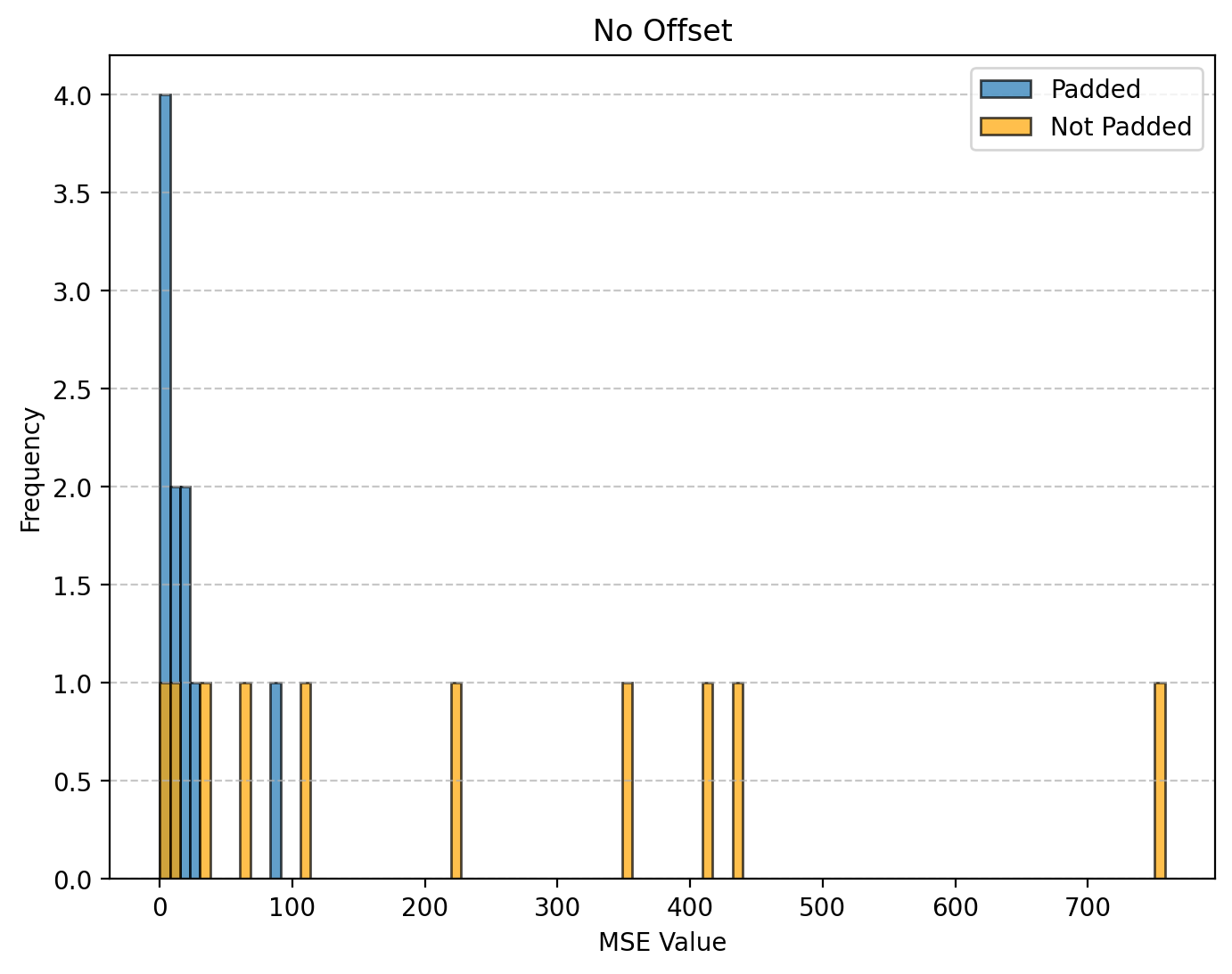} 
    \includegraphics[width=.48\textwidth]{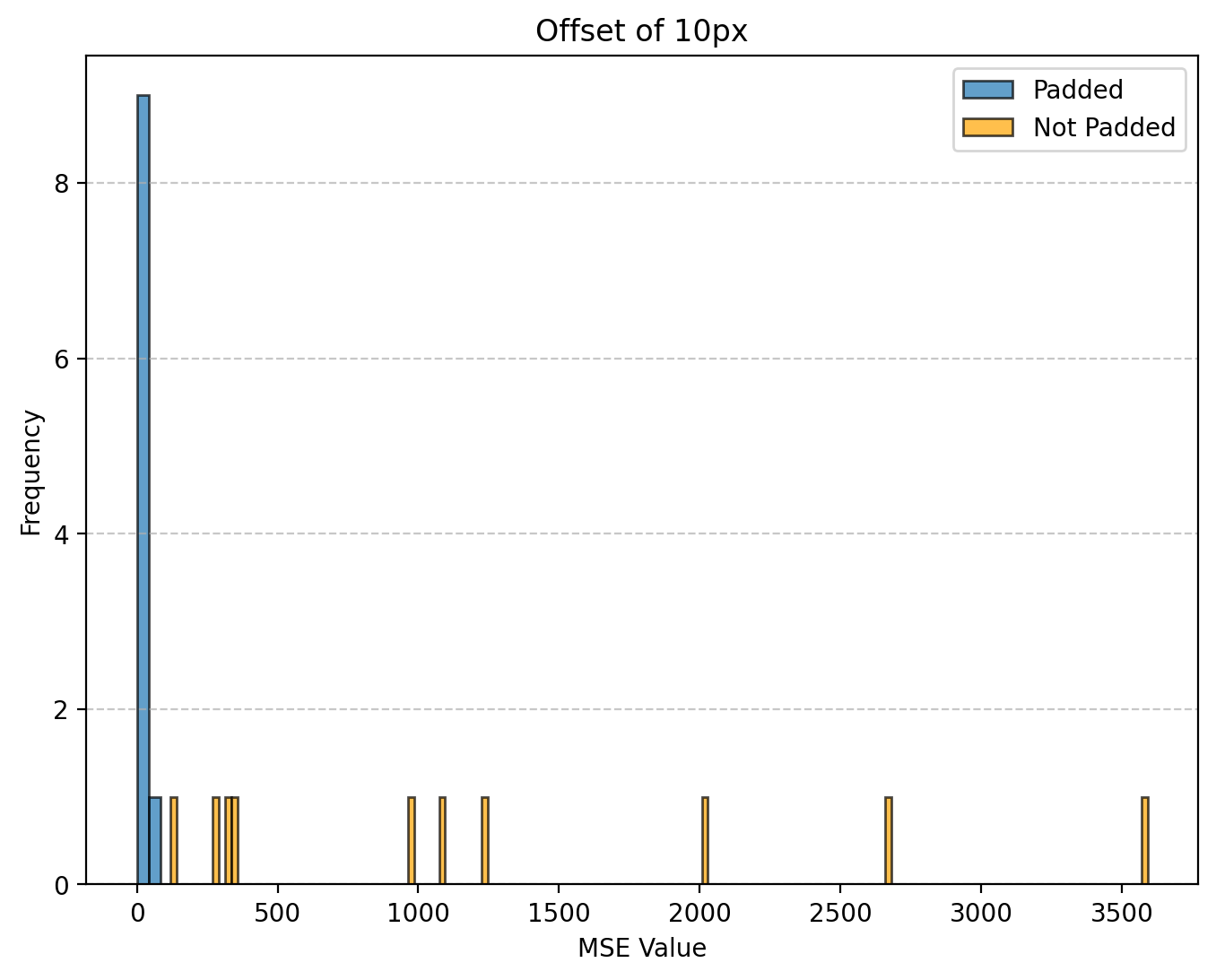} 
    \caption{Histograms of MSEs from 10 padded and 10 not padded images. The top graph is generated by setting the dividing line offset to 0, iterating the dividing line from the image boundary, and for the bottom graph, the dividing line offset is 10 pixels out from the boundary. As shown, when the offset is 0, the MSEs of the padded and not padded images are intersecting, disallowing for a clear threshold to differentiate between the two classes. In contrast, starting 10 pixels out disregards the edge cases where the pixels at the border have no difference, differentiating the MSEs between padded and unpadded images.}
    \label{fig:StartingPoint}
\end{figure}

We use a parameter of the 10th pixel as the dividing line offset for the threshold estimates.
A sample of MSEs from 10 images of unpadded and padded images is shown in Figure \ref{fig:StartingPoint} to illustrate the impact of this starting point hyperparameter. 

For the first iterative threshold method described in Section 3.1, we set the MSE-threshold to 70 and iterate it by 5 until 180. We calculate the precision and recall at each threshold value using the training sample. The start and end points of the threshold iteration were chosen based on where the precision or recall started to fall off. 
The threshold with the highest precision and recall is chosen as the optimal threshold. The second method takes the minimum MSEs from all images in the dataset and applies Otsu's method to estimate an optimal threshold. We evaluated this threshold with the precision and recall on the training sample. 


\begin{figure}
    \centering
    \includegraphics[width=.45\textwidth]{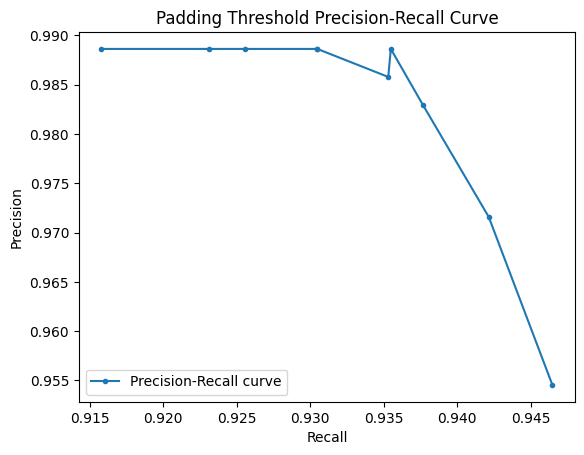} 
    \caption{Precision-recall curve of the accuracy of various thresholds in the task of differentiating the MSEs of padded and not-padded images.}
    \label{fig:PR}
\end{figure}

From the first method, we found that the best MSE-threshold is 110. The resulting precision-recall curve from the first method is shown in Figure \ref{fig:PR}

For the second method, the MSE-threshold given by Otsu's method is 1408. This discrepancy in thresholds is discussed later in the Discussion section. The precision and recall of both methods are provided in Table \ref{tab:thresholdResults}.


\subsection{Threshold Evaluation}
\begin{figure}
    \centering
    \includegraphics[width=.24\textwidth]{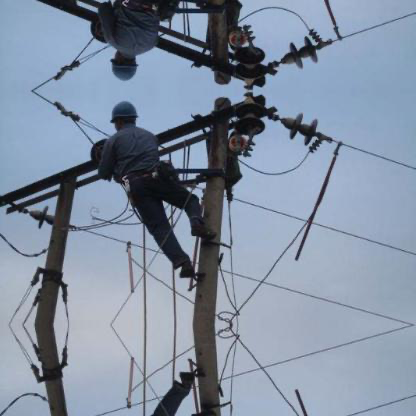} \includegraphics[width=.24\textwidth]{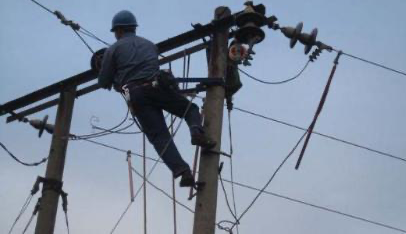} 
    \caption{Example of effectively removing the reflective padding of an image using the proposed algorithm and the best threshold obtained in the results. The first image is the original, and the second is the unpadded version.}
    \label{fig:paddingRemoval}
\end{figure}

Utilizing the best obtained MSE-threshold of 110, we apply the complete unpadding algorithm to the SHEL5k dataset, cropping the images and annotations accordingly to create a new dataset. 
We note the removed padding of the image in Figure \ref{fig:paddingRemoval} as a qualitative example.

We test for an improved model performance, specifically foundation vision language model performance in the task of zero-shot object detection \cite{minderer2024scaling, bansal2018zero, greer2025language, huang2024training, ghita2024activeanno3d, zhao2022exploiting} to benchmark this method without having variability in the training.

The model we utilize is OWLv2 \cite{minderer2024scaling}, and we use the cascaded detection strategy described by Choi and Greer \cite{choi2024construction}, specifically detecting hard hats inside of the bounding boxes of persons to automatically associate the two classes. For the prompts, we use ‘person,’ ‘helmet,’ and ‘hard hat.’

\begin{figure}
    \centering
    \includegraphics[width=.48\textwidth]{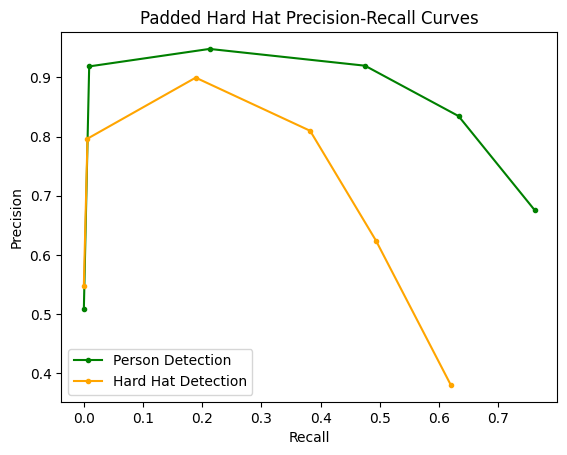} \includegraphics[width=.48\textwidth]{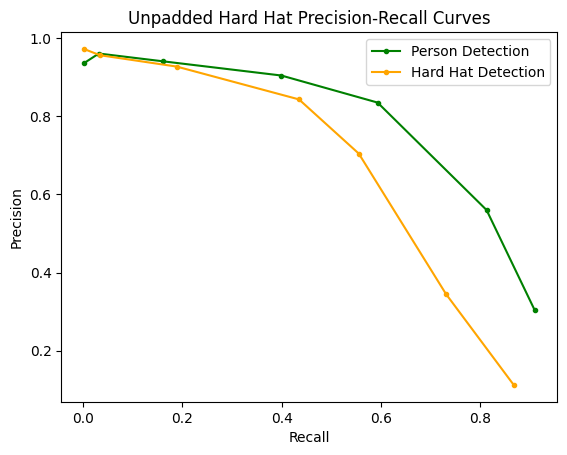} 
    \caption{Precision-recall curves of OWLv2's performance on the detection of hard hats and persons in the SHEL5k dataset. The first graph is the performance on the original dataset, and the second is after the images were unpadded. The first graph decreases faster and has an abnormally low starting point, suggesting that insufficient relevant detections were made at high thresholds, illustrating worse performance. In contrast, the second graph is concave down with a lower magnitude of the slope, having higher precision and recall throughout, demonstrating better performance. }
    \label{fig:testPRs}
\end{figure}

The model's performance in detecting hard hats and persons on both the original and unpadded datasets is provided in Table \ref{tab:OWLv2Result}, evaluated by average precision. The comparison in precision-recall curves is presented in Figure \ref{fig:testPRs}. There was a clear increase in performance after unpadding the data, as the removal of padding artifacts reduced misinterpretations of distorted persons by OWLv2.

\begin{table}[]
    \centering
    
    \begin{tabular}{c|c|c}
        Method & Precision & Recall \\
        \hline \hline
        Threshold Value Iteration & 0.9886 & 0.9355 \\
        \hline
        Otsu's Thresholding Method & 0.9915 & 0.6463 \\
    \end{tabular}
    \caption{Comparative evaluation of threshold methods through precision and recall}\label{tab:thresholdResults}
\end{table}

\begin{table}[]
    \centering
    
    \begin{tabular}{c|c|c}
        Dataset & Hard Hat (AP) & Person (AP) \\
        \hline \hline
        Original & 0.4672 & 0.6767 \\
        \hline
        Unpadded & 0.6115 & 0.7348 \\
    \end{tabular}
    \caption{Comparative evaluation of OWLv2's performance on original and processed dataset through average precision (AP)}\label{tab:OWLv2Result}
\end{table}


\section{Discussion}

The threshold estimation method utilizing Otsu's method is observed to have lower accuracy compared to iteratively testing the threshold as demonstrated in Table \ref{tab:thresholdResults}. This behavior is due to the MSEs and Otsu thresholds of the unpadded and padded images not having clear peaks and not being strongly separable. As shown in Figure \ref{fig:StartingPoint}, the MSEs of the unpadded images are dispersed, disallowing Otsu's method to find a clear point for the threshold. Therefore, manually testing the threshold based on observed patterns in the MSEs would be more accurate. This causes the large discrepancy of the thresholds from the first and second methods. 

\begin{figure}
    \centering
    \includegraphics[width=.48\textwidth]{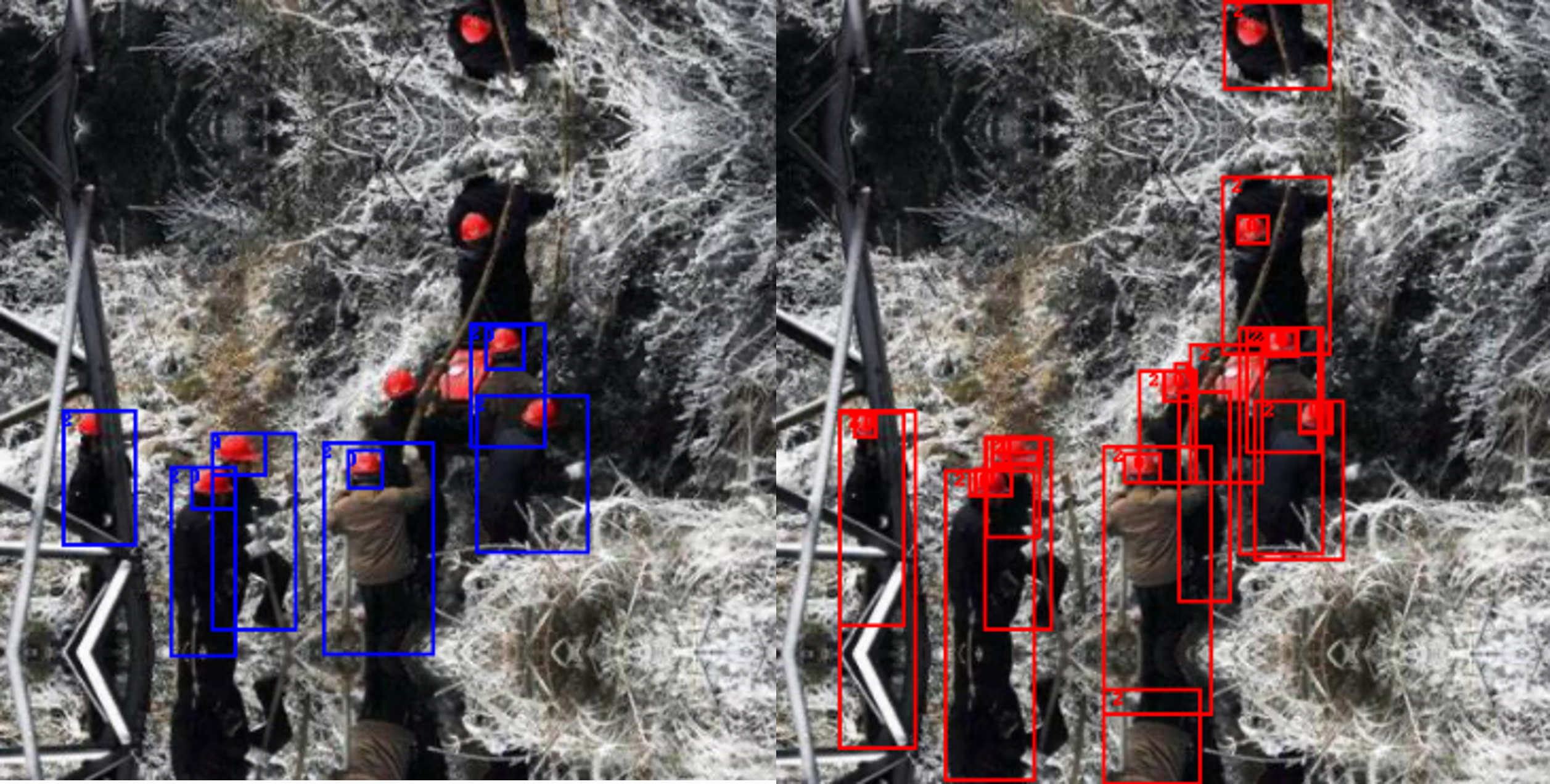} 
    \caption{The left image has drawn bounding boxes of the ground truth, showing the missing annotations in the padded regions. The right has drawn bounding boxes of the OWLv2 detections, showing the imprecise bounding boxes in the bottom padding.}
    \label{fig:MissedDetection}
\end{figure}

Our threshold evaluation showed improved performance after removing padding, as it reduced ambiguities that could mislead the model. Additionally, the original dataset's annotations for the mirrored padding were inconsistent, as often the entire person was not presented in the padding. This contributed to the model's false negatives, contributing to an inaccurate representation of OWLv2's performance. Figure \ref{fig:MissedDetection} demonstrates both the missing ground truth annotations of the padded area as well as OWLv2's imprecise detections in the symmetrically padded region. By removing the padding, our approach eliminates these ambiguous regions of missing annotations and distorted objects, leading to more accurate detections.

\section{Concluding Remarks}

In this research, we proposed an algorithm to remove noisy artificially padded mirrored areas from images, utilizing a minimization of the MSE with potential borders of the original image. Our method has achieved robust performance, effectively identifying the transition between mirrored and non-mirrored regions. This led to a significant increase in performance on the task of zero-shot detection as we allow for more precise and reliable predictions and evaluations, with an increase from 0.47 to 0.61 in OWLv2's average precision for hard hat detection and 0.68 to 0.73 in person detection.

Beyond artifact removal, ensuring that data augmentation techniques produce realistic transformations is an alternative solution to this reflective padding issue.  
Khalifa et al. \cite{khalifa2022comprehensive} demonstrate this possibility as they describe that recent advances in augmentation now include complex strategies such as adversarial training, neural style transfer, and synthetic data generation rather than simple geometric transformations. By leveraging neural networks, they look to avoid augmentations that inaccurately represent real-world scenarios. 

In light of recent trends in massive dataset augmentation by amalgamation of multi-source datasets to solve foundational learning tasks, we propose that the theme of finding unified visual qualities will be applicable in further use cases. For example, combined surveillance-related datasets may have combinations of both fisheye \cite{gochoo2023fisheye8k, tezcan2022wepdtof} and non-fisheye lenses \cite{li2018richly, oh2011large}, and detecting and correcting for these distortions is important for learning to detect visual patterns \cite{yang2023large, morris2008survey}. The ability to recognize and correct for fundamentally disagreeing patterns between datasets will be important to rectify by restoring images before network training.

As large-scale machine learning continues to stay prevalent, ensuring high-quality, naturalistic data remains essential for enhancing generalization and real-world applicability across diverse applications.


{\small
\bibliographystyle{ieeenat_fullname}
\bibliography{refs}

\begin{thebibliography}{34}
\providecommand{\natexlab}[1]{#1}
\providecommand{\url}[1]{\texttt{#1}}
\expandafter\ifx\csname urlstyle\endcsname\relax
  \providecommand{\doi}[1]{doi: #1}\else
  \providecommand{\doi}{doi: \begingroup \urlstyle{rm}\Url}\fi

\bibitem[Bansal et~al.(2018)Bansal, Sikka, Sharma, Chellappa, and Divakaran]{bansal2018zero}
Ankan Bansal, Karan Sikka, Gaurav Sharma, Rama Chellappa, and Ajay Divakaran.
\newblock Zero-shot object detection.
\newblock In \emph{Proceedings of the European conference on computer vision (ECCV)}, pages 384--400, 2018.

\bibitem[China(2022)]{hard-hat-workers_dataset}
Northeastern~University China.
\newblock Hard hat workers dataset.
\newblock \url{ https://universe.roboflow.com/joseph-nelson/hard-hat-workers }, 2022.
\newblock visited on 2024-06-24.

\bibitem[Chlap et~al.(2021)Chlap, Min, Vandenberg, Dowling, Holloway, and Haworth]{chlap2021review}
Phillip Chlap, Hang Min, Nym Vandenberg, Jason Dowling, Lois Holloway, and Annette Haworth.
\newblock A review of medical image data augmentation techniques for deep learning applications.
\newblock \emph{Journal of Medical Imaging and Radiation Oncology}, 65\penalty0 (5):\penalty0 545--563, 2021.

\bibitem[Choi and Greer(2024)]{choi2024construction}
Lucas Choi and Ross Greer.
\newblock Evaluating cascaded methods of vision-language models for zero-shot detection and association of hardhats for increased construction safety, 2024.

\bibitem[Das(2025)]{subhajeet_das_2025}
Subhajeet Das.
\newblock Iq-oth/nccd lung cancer dataset (augmented), 2025.

\bibitem[Elgendi et~al.(2021)Elgendi, Nasir, Tang, Smith, Grenier, Batte, Spieler, Leslie, Menon, Fletcher, et~al.]{elgendi2021effectiveness}
Mohamed Elgendi, Muhammad~Umer Nasir, Qunfeng Tang, David Smith, John-Paul Grenier, Catherine Batte, Bradley Spieler, William~Donald Leslie, Carlo Menon, Richard~Ribbon Fletcher, et~al.
\newblock The effectiveness of image augmentation in deep learning networks for detecting covid-19: A geometric transformation perspective.
\newblock \emph{Frontiers in Medicine}, 8:\penalty0 629134, 2021.

\bibitem[Fernández(2025)]{heartdataset}
Miguel Fernández.
\newblock Markov transition field images of heart beats, 2025.

\bibitem[Geirhos et~al.(2022)Geirhos, Narayanappa, Mitzkus, Thieringer, Bethge, Wichmann, and Brendel]{geirhos2022bittersweet}
Robert Geirhos, Kantharaju Narayanappa, Benjamin Mitzkus, Tizian Thieringer, Matthias Bethge, Felix~A Wichmann, and Wieland Brendel.
\newblock The bittersweet lesson: data-rich models narrow the behavioural gap to human vision.
\newblock \emph{Journal of Vision}, 22\penalty0 (14):\penalty0 3273--3273, 2022.

\bibitem[Ghita et~al.(2024)Ghita, Antoniussen, Zimmer, Greer, Cre{\ss}, M{\o}gelmose, Trivedi, and Knoll]{ghita2024activeanno3d}
Ahmed Ghita, Bj{\o}rk Antoniussen, Walter Zimmer, Ross Greer, Christian Cre{\ss}, Andreas M{\o}gelmose, Mohan~M Trivedi, and Alois~C Knoll.
\newblock Activeanno3d-an active learning framework for multi-modal 3d object detection.
\newblock In \emph{2024 IEEE Intelligent Vehicles Symposium (IV)}, pages 1699--1706. IEEE, 2024.

\bibitem[Gochoo et~al.(2023)Gochoo, Otgonbold, Ganbold, Hsieh, Chang, Chen, Dorj, Al~Jassmi, Batnasan, Alnajjar, et~al.]{gochoo2023fisheye8k}
Munkhjargal Gochoo, Munkh-Erdene Otgonbold, Erkhembayar Ganbold, Jun-Wei Hsieh, Ming-Ching Chang, Ping-Yang Chen, Byambaa Dorj, Hamad Al~Jassmi, Ganzorig Batnasan, Fady Alnajjar, et~al.
\newblock Fisheye8k: A benchmark and dataset for fisheye camera object detection.
\newblock In \emph{Proceedings of the IEEE/CVF conference on computer vision and pattern recognition}, pages 5305--5313, 2023.

\bibitem[Gohil and Gurjar(2025)]{licenseplatedataset}
Ronak Gohil and Amit Gurjar.
\newblock License plate dataset, 2025.

\bibitem[Greer et~al.(2024)Greer, Gopalkrishnan, Keskar, and Trivedi]{greer2024patterns}
Ross Greer, Akshay Gopalkrishnan, Maitrayee Keskar, and Mohan~M Trivedi.
\newblock Patterns of vehicle lights: Addressing complexities of camera-based vehicle light datasets and metrics.
\newblock \emph{Pattern Recognition Letters}, 178:\penalty0 209--215, 2024.

\bibitem[Greer et~al.(2025)Greer, Antoniussen, M{\o}gelmose, and Trivedi]{greer2025language}
Ross Greer, Bj{\o}rk Antoniussen, Andreas M{\o}gelmose, and Mohan Trivedi.
\newblock Language-driven active learning for diverse open-set 3d object detection.
\newblock In \emph{Proceedings of the Winter Conference on Applications of Computer Vision}, pages 980--988, 2025.

\bibitem[Huang et~al.(2024)Huang, Zheng, Wang, Xia, Pavone, and Huang]{huang2024training}
Rui Huang, Henry Zheng, Yan Wang, Zhuofan Xia, Marco Pavone, and Gao Huang.
\newblock Training an open-vocabulary monocular 3d detection model without 3d data.
\newblock \emph{Advances in Neural Information Processing Systems}, 37:\penalty0 72145--72169, 2024.

\bibitem[Keskar et~al.(2025)Keskar, Greer, Gopalkrishnan, Deo, and Trivedi]{keskar2025lights}
Maitrayee Keskar, Ross Greer, Akshay Gopalkrishnan, Nachiket Deo, and Mohan Trivedi.
\newblock Lights as points: Learning to look at vehicle substructures with anchor-free object detection.
\newblock \emph{Robotics and Automation Letters}, 2025.

\bibitem[Khalifa et~al.(2022)Khalifa, Loey, and Mirjalili]{khalifa2022comprehensive}
Nour~Eldeen Khalifa, Mohamed Loey, and Seyedali Mirjalili.
\newblock A comprehensive survey of recent trends in deep learning for digital images augmentation.
\newblock \emph{Artificial Intelligence Review}, 55\penalty0 (3):\penalty0 2351--2377, 2022.

\bibitem[Klug and Heckel()]{klugscaling}
Tobit Klug and Reinhard Heckel.
\newblock Scaling laws for deep learning based image reconstruction.
\newblock In \emph{The Eleventh International Conference on Learning Representations}.

\bibitem[Li et~al.(2018)Li, Zhang, Chen, and Huang]{li2018richly}
Dangwei Li, Zhang Zhang, Xiaotang Chen, and Kaiqi Huang.
\newblock A richly annotated pedestrian dataset for person retrieval in real surveillance scenarios.
\newblock \emph{IEEE transactions on image processing}, 28\penalty0 (4):\penalty0 1575--1590, 2018.

\bibitem[Mark et~al.(1982)Mark, Schluter, Moody, Devlin, and Chernoff]{mark1982annotated}
RG Mark, PS Schluter, G Moody, P Devlin, and D Chernoff.
\newblock An annotated ecg database for evaluating arrhythmia detectors.
\newblock In \emph{IEEE Transactions on Biomedical Engineering}, pages 600--600. IEEE-INST ELECTRICAL ELECTRONICS ENGINEERS INC 345 E 47TH ST, NEW YORK, NY~…, 1982.

\bibitem[Miko{\l}ajczyk and Grochowski(2018)]{mikolajczyk2018data}
Agnieszka Miko{\l}ajczyk and Micha{\l} Grochowski.
\newblock Data augmentation for improving deep learning in image classification problem.
\newblock In \emph{2018 international interdisciplinary PhD workshop (IIPhDW)}, pages 117--122. IEEE, 2018.

\bibitem[Minderer et~al.(2024)Minderer, Gritsenko, and Houlsby]{minderer2024scaling}
Matthias Minderer, Alexey Gritsenko, and Neil Houlsby.
\newblock Scaling open-vocabulary object detection.
\newblock \emph{Advances in Neural Information Processing Systems}, 36, 2024.

\bibitem[Morris and Trivedi(2008)]{morris2008survey}
Brendan~Tran Morris and Mohan~Manubhai Trivedi.
\newblock A survey of vision-based trajectory learning and analysis for surveillance.
\newblock \emph{IEEE transactions on circuits and systems for video technology}, 18\penalty0 (8):\penalty0 1114--1127, 2008.

\bibitem[Nagaraju et~al.(2022)Nagaraju, Chawla, and Kumar]{nagaraju2022performance}
Mamillapally Nagaraju, Priyanka Chawla, and Neeraj Kumar.
\newblock Performance improvement of deep learning models using image augmentation techniques.
\newblock \emph{Multimedia Tools and Applications}, 81\penalty0 (7):\penalty0 9177--9200, 2022.

\bibitem[Oh et~al.(2011)Oh, Hoogs, Perera, Cuntoor, Chen, Lee, Mukherjee, Aggarwal, Lee, Davis, et~al.]{oh2011large}
Sangmin Oh, Anthony Hoogs, Amitha Perera, Naresh Cuntoor, Chia-Chih Chen, Jong~Taek Lee, Saurajit Mukherjee, Jake~K Aggarwal, Hyungtae Lee, Larry Davis, et~al.
\newblock A large-scale benchmark dataset for event recognition in surveillance video.
\newblock In \emph{CVPR 2011}, pages 3153--3160. IEEE, 2011.

\bibitem[Otgonbold et~al.(2022)Otgonbold, Gochoo, Alnajjar, Ali, Tan, Hsieh, and Chen]{otgonbold2022shel5k}
Munkh-Erdene Otgonbold, Munkhjargal Gochoo, Fady Alnajjar, Luqman Ali, Tan-Hsu Tan, Jun-Wei Hsieh, and Ping-Yang Chen.
\newblock Shel5k: An extended dataset and benchmarking for safety helmet detection.
\newblock \emph{Sensors}, 22\penalty0 (6):\penalty0 2315, 2022.

\bibitem[Otsu et~al.(1975)]{otsu1975threshold}
Nobuyuki Otsu et~al.
\newblock A threshold selection method from gray-level histograms.
\newblock \emph{Automatica}, 11\penalty0 (285-296):\penalty0 23--27, 1975.

\bibitem[Schulz et~al.(2020)Schulz, Yeo, Vogelstein, Mourao-Miranada, Kather, Kording, Richards, and Bzdok]{schulz2020different}
Marc-Andre Schulz, BT~Thomas Yeo, Joshua~T Vogelstein, Janaina Mourao-Miranada, Jakob~N Kather, Konrad Kording, Blake Richards, and Danilo Bzdok.
\newblock Different scaling of linear models and deep learning in ukbiobank brain images versus machine-learning datasets.
\newblock \emph{Nature communications}, 11\penalty0 (1):\penalty0 4238, 2020.

\bibitem[Shorten and Khoshgoftaar(2019)]{shorten2019survey}
Connor Shorten and Taghi~M Khoshgoftaar.
\newblock A survey on image data augmentation for deep learning.
\newblock \emph{Journal of big data}, 6\penalty0 (1):\penalty0 1--48, 2019.

\bibitem[Sun et~al.(2017)Sun, Shrivastava, Singh, and Gupta]{sun2017revisiting}
Chen Sun, Abhinav Shrivastava, Saurabh Singh, and Abhinav Gupta.
\newblock Revisiting unreasonable effectiveness of data in deep learning era.
\newblock In \emph{Proceedings of the IEEE international conference on computer vision}, pages 843--852, 2017.

\bibitem[Tezcan et~al.(2022)Tezcan, Duan, Cokbas, Ishwar, and Konrad]{tezcan2022wepdtof}
Ozan Tezcan, Zhihao Duan, Mertcan Cokbas, Prakash Ishwar, and Janusz Konrad.
\newblock Wepdtof: A dataset and benchmark algorithms for in-the-wild people detection and tracking from overhead fisheye cameras.
\newblock In \emph{Proceedings of the IEEE/CVF winter conference on applications of computer vision}, pages 503--512, 2022.

\bibitem[Xu et~al.(2023)Xu, Yoon, Fuentes, and Park]{xu2023comprehensive}
Mingle Xu, Sook Yoon, Alvaro Fuentes, and Dong~Sun Park.
\newblock A comprehensive survey of image augmentation techniques for deep learning.
\newblock \emph{Pattern Recognition}, 137:\penalty0 109347, 2023.

\bibitem[Yang et~al.(2023)Yang, Li, Xin, Sun, Song, and Wang]{yang2023large}
Lu Yang, Liulei Li, Xueshi Xin, Yifan Sun, Qing Song, and Wenguan Wang.
\newblock Large-scale person detection and localization using overhead fisheye cameras.
\newblock In \emph{Proceedings of the IEEE/CVF International Conference on Computer Vision}, pages 19961--19971, 2023.

\bibitem[Zhang(2025)]{maskdataset}
Edward Zhang.
\newblock Face mask detection, 2025.

\bibitem[Zhao et~al.(2022)Zhao, Zhang, Schulter, Zhao, Vijay~Kumar, Stathopoulos, Chandraker, and Metaxas]{zhao2022exploiting}
Shiyu Zhao, Zhixing Zhang, Samuel Schulter, Long Zhao, BG Vijay~Kumar, Anastasis Stathopoulos, Manmohan Chandraker, and Dimitris~N Metaxas.
\newblock Exploiting unlabeled data with vision and language models for object detection.
\newblock In \emph{European conference on computer vision}, pages 159--175. Springer, 2022.

\end{thebibliography}
}


\end{document}